\documentclass[journal]{IEEEtran}
\usepackage{amsmath,amsfonts}

\usepackage{array}
\usepackage[caption=false,font=normalsize,labelfont=sf,textfont=sf]{subfig}
\usepackage{textcomp}
\usepackage{stfloats}
\usepackage{url}
\usepackage{verbatim}
\usepackage{graphicx}
\usepackage{cite}

\usepackage{textcomp}
\usepackage{stfloats}
\usepackage{url}
\usepackage{verbatim}
\usepackage{graphicx}
\usepackage{cite}

\usepackage{textcomp}
\usepackage{stfloats}
\usepackage{url}
\usepackage{verbatim}
\usepackage{graphicx}
\usepackage{cite}

\usepackage{makecell}
\usepackage{times}
\usepackage{epsfig}
\usepackage{xcolor}
\usepackage{setspace}
\usepackage{multirow}
\usepackage{array}
\usepackage{verbatim}
\usepackage{amsmath}
\usepackage[ruled,norelsize,vlined,linesnumbered]{algorithm2e}

\usepackage{arydshln}
\RequirePackage{xspace}
\RequirePackage{graphicx}
\RequirePackage{amsmath}
\RequirePackage{amssymb}
\RequirePackage{booktabs}

\makeatletter
\DeclareRobustCommand\onedot{\futurelet\@let@token\@onedot}
\def\@onedot{\ifx\@let@token.\else.\null\fi\xspace}

\def\etal{\emph{et al}\onedot}
\makeatother

\begin{document}

\title{ONNXPruner: ONNX-Based General Model Pruning Adapter}

\author{Dongdong~Ren,
        Wenbin~Li,
        Tianyu~Ding,
        Lei~Wang,
        Qi~Fan,
        Jing~Huo,
        Hongbing~Pan,
        and~Yang~Gao
        % <-this % stops a space
\thanks{Dongdong~Ren, Wenbin~Li, Qi~Fan, Jing~Huo, Yang~Gao are with the State Key Laboratory for Novel Software Technology, Nanjing University, China (e-mail: rdd@smail.nju.edu.cn, liwenbin@nju.edu.cn,  fanqics@gmail.com, huojing@nju.edu.cn, gaoy@nju.edu.cn).}
\thanks{Tianyu~Ding is with the Microsoft, Redmond, WA 98052, USA (e-mail: tianyuding@microsoft.com).}
\thanks{Lei~Wang is with the School of Computing and Information Technology, University of Wollongong, NSW 2522, Australia (e-mail: leiw@uow.edu.au).}
\thanks{Hongbing~Pan is with the School of Electronic Science and Engineering, Nanjing University, China (e-mail: phb@nju.edu.cn).}}

\markboth{Journal of \LaTeX\ Class Files,~Vol.~14, No.~8, August~2021}%
{Shell \MakeLowercase{\textit{et al.}}: A Sample Article Using IEEEtran.cls for IEEE Journals}

\IEEEpubid{0000--0000/00\$00.00~\copyright~2021 IEEE}

\maketitle

\begin{abstract}
Recent advancements in model pruning have focused on developing new algorithms and improving upon benchmarks. However, the practical application of these algorithms across various models and platforms remains a significant challenge. To address this challenge, we propose ONNXPruner, a versatile  pruning adapter designed for the ONNX format models. ONNXPruner streamlines the adaptation process across diverse deep learning frameworks and hardware platforms. A novel aspect of ONNXPruner is its use of node association trees, which automatically adapt to various model architectures. These trees clarify the structural relationships between nodes, guiding the pruning process, particularly highlighting the impact on interconnected nodes. Furthermore, we introduce a tree-level evaluation method. By leveraging node association trees, this method allows for a comprehensive analysis beyond traditional single-node evaluations, enhancing pruning performance without the need for extra operations. Experiments across multiple models and datasets confirm ONNXPruner's strong adaptability and increased efficacy. Our work aims to advance the practical application of model pruning.
\end{abstract}  
\begin{IEEEkeywords}
General model pruning, ONNX, tree-level evaluation, deep neural network.
\end{IEEEkeywords}

\section{Introduction}
\label{sec:intro}
\IEEEPARstart{I}{n} recent years, deep learning has achieved remarkable successes in many different fields. However, the deployment of deep neural network (DNN) models in practical applications is hindered by their substantial model size and intensive computational demands. To mitigate these challenges and accelerate inference speed while maintaining performance, researchers have paid much attention to DNN model compression, particularly for edge computing~\cite{fang2023depgraph,han2015learning,liu2021group,jing2021meta, kuchaiev2019nemo,yao2021wenet, 8416559, 9457173}.

Among the various model compression techniques, network pruning has been becoming a popular method, known for its efficiency and portability~\cite{wang2021convolutional,chao2020directional}. Generally, network pruning methods fall into two categories~\cite{10330640}. One is unstructured pruning~\cite{han2015learning,kwon2020structured,chen2021orthant,chen2020neural}, which creates sparse models by zeroing out weights of less important filters, but often requires specialized hardware or software to effectively handle sparse matrices. The other is structured pruning~\cite{liu2017learning,li2016pruning,he2019filter,malach2020proving,chen2021only,chen2023otov}, which removes unimportant or redundant filters, offering a more hardware-agnostic approach, and thus has become more widely used~\cite{liu2021content}.

\begin{figure}[t]
	\centering
	\begin{minipage}[b]{0.9\linewidth}
		\includegraphics[width=1\linewidth]{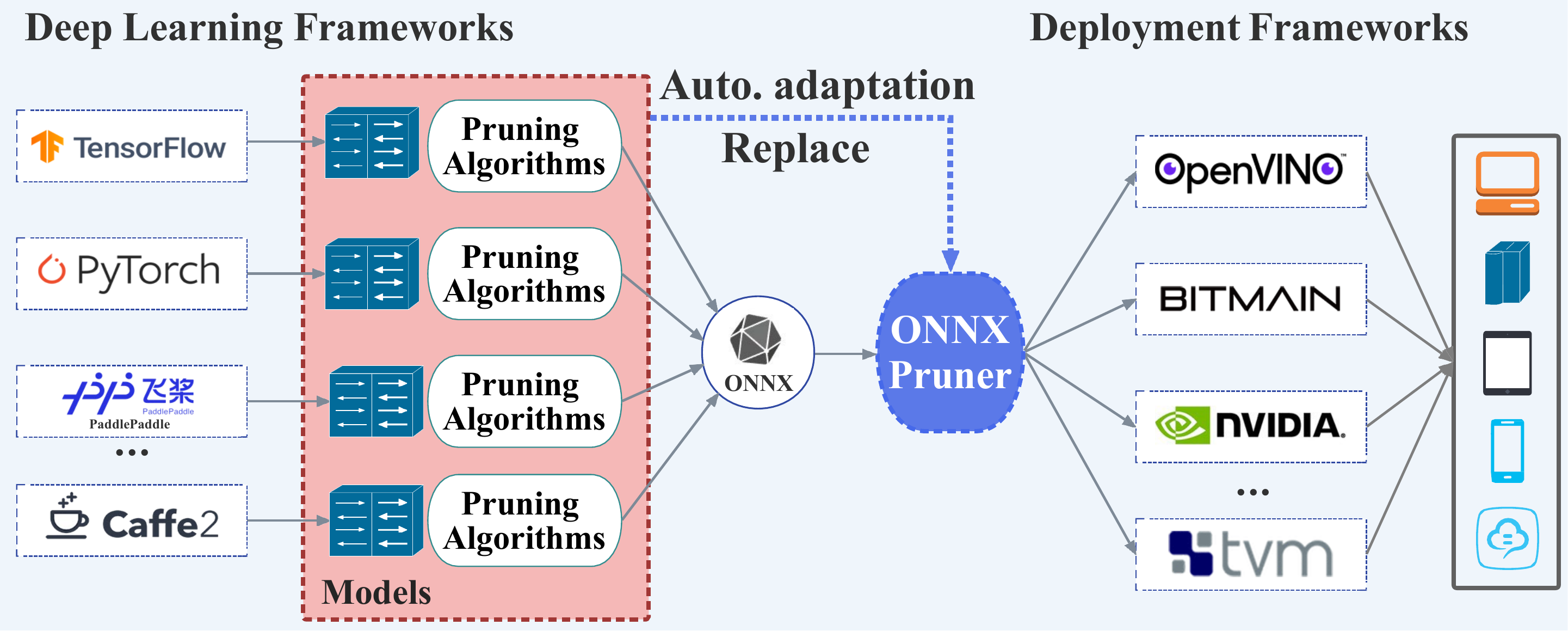}
	\end{minipage}
	\caption{Illustration of pruning algorithms in development and deployment. Existing pruning algorithms (red boxes) are tailored for specific development frameworks and necessitate manual adaptation for various model structures. Our work introduces ONNXPruner (blue box), a versatile model pruning adapter for ONNX format models, which provides automatic adaptation of pruning algorithms to models with diverse structures, in-depending on the development framework used.}
	\label{fig:1}
\end{figure}

\begin{figure*}[t]
	\centering
	\begin{minipage}[b]{1\linewidth}
		\includegraphics[width=1\linewidth]{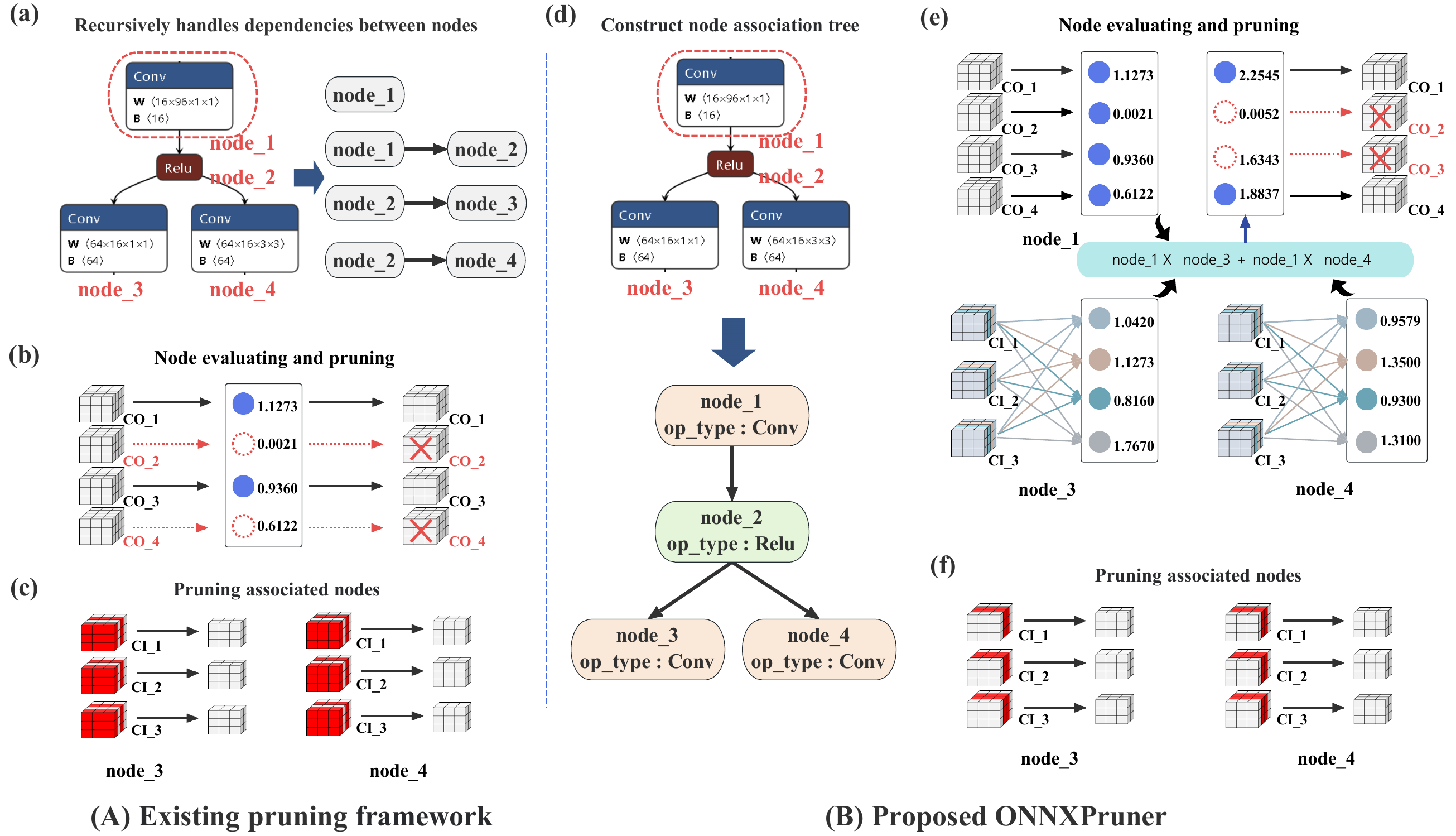}
	\end{minipage}
	\caption{Comparison between the existing pruning framework (A) and the proposed pruning framework (B). (a)~The current method~\cite{fang2023depgraph} recursively traverses each node to identify the associated nodes. (b)~Evaluates filters based solely on the pruned node for pruning decisions. (c)~Prunes associated nodes based on the evaluation in (b). (d)~The proposed ONNXPruner constructs node association trees to represent the relationships between a pruned node and its associated nodes, presenting these relationships hierarchically. (e)~Employs a tree-level pruning strategy, leveraging the node association tree to jointly evaluate and prune the filters of both the pruned node and associated nodes. (f)~Prunes associated nodes following the evaluation in (e). CO and CI represent the output and input channels of the Conv kernel, respectively.}
	\label{fig:2}
 %\vspace{-0.5cm}
\end{figure*}

However, existing model pruning techniques face two main challenges in their application to device environments. First, existing pruning algorithms are typically tied to specific deep learning frameworks and it is difficult to transfer them across different platforms. For instance, developers often have to recreate pruning algorithms for each framework (like Keras, PyTorch, PaddlePaddle, \textit{etc.}), convert them to ONNX (Open Neural Network Exchange) format~\cite{ONNX}, and then deploy on devices. This process is time-consuming and labor-intensive. Second, developers must tailor pruning procedures for different model structures. The complex internal connections in DNNs can lead to cascading effects when pruning nodes. Although some studies have attempted to address this issue by constructing chains of relationships between pruned and associated nodes~\cite{you2019gate,liu2021group,fang2023depgraph}, they generally lack an explicit model of these relationships, requiring additional components and operations for evaluation.
\IEEEpubidadjcol

Aiming at solving the above challenges, we propose a novel \textit{ONNX-based general model pruning adapter}, named ONNXPruner. Specifically, ONNXPruner leverages the ONNX framework to improve the interoperability of pruning algorithms across different application systems, as shown in Figure~\ref{fig:1}. ONNX, being a cross-platform deep learning model exchange format, facilitates easy conversion and deployment between various deep learning frameworks and hardware platforms. It is widely supported by numerous deep learning frameworks and hardware acceleration platforms. Thus, we initiate pruning research on ONNX models to enhance the adaptability of pruning algorithms in diverse applications. Crucially, to mitigate the cascading effects caused by pruning nodes, ONNXPruner constructs a node association tree for each pruned node within a model. This allows the pruning algorithm to automatically adjust to different model structures by clearly defining the relationships between a pruned node and its associated nodes. This feature enables the pruner to effectively track changes to associated nodes following the pruning of a node. Furthermore, while structured pruning typically involves the simultaneous removal of parameters from a pruned node and its associated nodes, existing pruning algorithms often only assess the pruned node, overlooking the impact of this action~\cite{filterspruning,SFP,Hranklin2020hrank}, as shown in Figure~\ref{fig:2}(a). To address this limitation, we design a \textit{tree-level pruning method} that utilizes node association trees to comprehensively evaluate channels, facilitating the removal of unimportant weights more effectively, as demonstrated in Figure~\ref{fig:2}(b). This approach efficiently manages complex node associations without the need for additional components or operations.

%To bridge these two gaps in the application of model pruning techniques to device applications, this paper proposes an ONNX based general model pruning adapter that focuses on addressing the above two issues.
%\textbf{Selecting the Open Neural Network Exchange format ONNX as the development framework.} Existing end-device deployment solutions commonly involve first converting the model to the ONNX format and then adapting it to deep learning compilers and hardware platforms. ONNX is a cross-platform deep learning model exchange format that can be easily converted and deployed between different deep learning frameworks and hardware platforms, and is supported by a variety of deep learning frameworks and hardware acceleration platforms.However, most pruning algorithms are aimed at a single deep learning framework, and need to be re-adapted when encountering models from different frameworks. In this work, we developed an ONNX based pruning framework,  which has deep learning framework independence, improving the deployment flexibility, as shown in Figure~\ref{fig:1}.\textbf{Constructing node association trees to make the pruning algorithm adaptive to various model structures.}We propose node association trees that explicitly construct the relationship between pruned nodes and associated nodes, accurately querying the relationship between the root node and each child. 

To validate the performance of ONNXPruner, we tailor multiple popular pruning algorithms as baselines within ONNXPruner, including $\ell_1$-norm~\cite{filterspruning}, $\ell_2$-norm~\cite{SFP}, and Hrank~\cite{Hranklin2020hrank}. ONNXPruner exhibits superior performance across a variety of model structures and tasks when compared to these baseline methods. To accommodate different model architectures, we establish a node attribute library for ONNXPruner.  This library categorizes the model nodes' operator types into four attributes: pruned, next-no-process, next-process, and stop-process, which are essential for constructing node association trees. The library currently encompasses over 35 types of DNN operators and has been validated on both CNNs and Transformer models, including AlexNet~\cite{krizhevsky2012imagenet}, SqueezeNet~\cite{iandola2016squeezenet}, VGG~\cite{simonyan2014very}, ResNet18~\cite{he2016deep}, FCN~\cite{FCNlong2015fully}, PSPNet~\cite{PSPzhao2017pyramid}, and ViT~\cite{dosovitskiy2020image}, among others.

In summary, we propose ONNXPruner, a versatile model pruning adapter designed to fit a wide range of model architectures through the interoperable ONNX format. The goal of ONNXPruner is not to invent new pruning algorithms but to provide a general-purpose pruning tool that allows application developers to effortlessly implement pruning algorithms. The main contributions of this work are as follows:
\begin{itemize}
\item To the best of our knowledge, this is the first attempt to develop a general pruning adapter for ONNX. This adapter significantly improves the interoperability of pruning algorithms across different device applications.
\item We introduce node association trees to clearly define the relationships between pruned nodes and their associated nodes. This innovation enables pruning adapters to effectively manage various model structures.
\item We develop a tree-level evaluation method utilizing node association trees. This approach allows for the assessment of diverse node connectivity structures without extra components and demonstrates superior performance.
\end{itemize}

\section{Related Work}
\label{sec:formatting}

\subsection{Model Conversion for Interoperability}
In deep learning application systems, interoperability refers to the ability of software to exchange algorithms and models efficiently~\cite{jajal2023analysis}. This software primarily encompasses development frameworks~\cite{paszke2019pytorch,abadi2016tensorflow,chen2015mxnet} and compilers~\cite{chen2018tvm,TensorRT,OpenVINO} for deep learning.  Given the vast array of development and deployment options, transferring algorithms between different frameworks is challenging. Moreover, compilers do not universally support models from all frameworks. Consequently, intermediaries like ONNX~\cite{ONNX} have become crucial for enhancing the interoperability of deep learning software, as depicted in Figure~\ref{fig:1}. ONNX uses the protocol buffers data structure and standardizes model representation through control flow in a graphical format, simplifying the conversion of models to the ONNX format. This facilitates the easy transfer of models across various frameworks, thus boosting interoperability.

In this context, we introduce a novel ONNX-based generic model pruning method. This approach, for the first time, circumvents the need for algorithmic porting between different frameworks. It allows for the seamless integration of pruned models with deep learning compilers, enhancing the interoperability of pruning algorithms from the development phase to deployment.

\subsection{Model Pruning Method}

Mainstream pruning methods fall into two categories: structural pruning~\cite{filterspruning,SFP,liu2021group,you2019gate,8485719} and unstructural pruning~\cite{dong2017learning, lee2019signal, park2020lookahead}. Unstructured pruning, which creates sparse matrices by zeroing out less important weights, often requires specialized hardware and software. Consequently, current research predominantly focuses on structured pruning. We now briefly review key studies in structured pruning, which is further divided based on the pruning timeline: before training, during training, and after training.

\textbf{Pruning before training} involves reducing the network size before the training process begins, utilizing concepts like the lottery hypothesis or sensitivity analysis to identify a lightweight sub-network early on~\cite{Namhoon:ICLR2019, Chaoqi:ICLR2020,Pau:ICLR2021}. This approach can expedite training by making the network sparse from the start~\cite{Jonathan:ICML2020}, but its stability is generally low, potentially leading to significant performance variability in the pruned model.

\textbf{Pruning during training} employs a cyclical process of training, pruning, and evaluating to minimize performance loss. One common method~\cite{Yang:IJCAI2018} sets filters with low $\ell_n$-norm values to zero at each training epoch, removing those zero-value filters upon training completion. Other strategies incorporate reinforcement learning or neural architecture search to dynamically prune filters during the training process~\cite{Tao:ICLR2020,10265172}. These methods help mitigate the adverse effects of filter removal but extend the training duration and necessitate specific adjustments during training.

\textbf{Pruning after training} follows a train-prune-finetune approach, where a large model is first trimmed down based on certain criteria and then refined to recover performance. The crucial aspect here is defining the importance of filters, typically through the calculation of their $\ell_n$-norm~\cite{filterspruning, SFP}. This assumption holds that filters with smaller $\ell_n$-norms are less crucial for the network. Such methods are straightforward, requiring no extra fine-tuning computations and are commonly employed in various applications~\cite{Yuchen:CVPR2021,Lewei:CVPR2021}.

\begin{table}[t]\footnotesize
\renewcommand{\arraystretch}{1.5}
  \centering
  \caption{The ONNX conversion process includes constructing node graphs, transforming nodes for equivalence, optimizing nodes, serializing the model, checking for correctness, and verifying performance.}
  \begin{tabular}{p{0.2cm}<{}p{1.9cm}<{}p{5cm}<{}}
    \toprule
    Step & Operation & Definition\\
    \midrule
    S1 & Load Model & Loading models of native frameworks\\
    S2 & Representation & ONNX graph tracing dynamic graph\\
    S3 & Node conversion & Graph nodes replaced by ONNX equivalents\\
    S4 & Optimization & Elimination of redundant nodes\\
    S5 & Export & Model serialized into protocol buffer\\
    S6 & Check & Syntactic checks and semantic checks \\
    S7 & Validate & Test whether the inference result is correct\\
    \bottomrule
  \end{tabular}
  
  \label{tab:onnxcoverter}
\end{table}

\subsection{General Model Pruning Method}

In deep neural networks, the interconnections between neural nodes are generally intricate~\cite{he2016deep,huang2017densely}. When a node is pruned or reduced, its associated nodes must be adjusted to maintain normal inference functionality. For instance, in structural pruning, pruning some filters in a node necessitates the removal of corresponding input channels in the filters of the subsequent layer. In simpler models, it's feasible to manually determine which layers to prune, a common approach in many existing methods~\cite{he2017channel, filterspruning}. However, this becomes challenging with complex model architectures, especially those featuring intricate residual connections and layer operations.

Recent efforts have aimed to unravel these complex layer relationships. Liu \etal.~\cite{liu2021group} and You \etal.~\cite{you2019gate} propose pruning algorithms for residual structures, employing a uniform mask to prune the two input nodes of the add operator within residual blocks. Similarly, Fang \etal.~\cite{fang2023depgraph} and Chen \etal.~\cite{chen2023otov} introduce automatic pruning framework that models relationships between layers and groups dependent nodes. For residual structures, thanks to their identified groups, both of them enable the pruning of the two input nodes of the add operator. 
In addition, we noticed that a work around the same time as ours also pruned the ONNX model~\cite{wang2024structurally}, but it also introduced masks to handle complex connections.
Nonetheless, these methods have several limitations: (1) They require adding new structures to the model or retraining it to handle complex node couplings, like residual connections; (2) Their approach of recursively grouping associated nodes lacks a clear construction of relationships within the entire group; (3) These methods are developed primarily using PyTorch, resulting in limited interoperability across different deep learning frameworks. 

To address these challenges, we propose an ONNX-based general model pruning framework in this paper. It clearly establishes a node correlation tree between a pruned node and its associated nodes, without adding extra components or necessitating re-training.

\begin{table*}[!t]%\small
\renewcommand{\arraystretch}{1.5}
  \centering
  \caption{Node attribute library. We categorize node attributes according to their operator types to facilitate the construction of node association trees. We distinguish operators into four node attributes: pruned refers to a model node requiring pruning, next-no-process for associated nodes that do not need further processing, next-process for those that do, and stop-process for nodes where the search for further children ends post-processing.}
  
  \begin{tabular}{p{2.5cm}<{}p{10cm}<{\centering}p{0.8cm}<{}}
  
    \toprule
    Node attribute & Operator type & Tree\\
    \midrule
    pruned\vspace{0.1cm} & Conv, ConvTranspose, Gemm, MatMul, Mul \vspace{0.1cm}& root\vspace{0.1cm}\\
    next-no-process \vspace{0.1cm}&  \makecell{Relu, Sigmoid, Softmax, Tanh, MaxPool, AveragePoo, Flatten \\ GlobalAveragePool, Pad, Reshape, Transpose, ReduceMean\\ ReduceMax, Pow, Sqrt, Erf, Unsqueeze, Resize, Slice, Cast }\vspace{0.1cm}& child\vspace{0.1cm}\\
    next-process\vspace{0.1cm} & Add, Concat, BatchNormalization, Sub, Div, Gather\vspace{0.1cm} & child\vspace{0.1cm}\\
    stop-process & Conv, ConvTranspose, Gemm, MatMul, Mul & leaf\\
    \bottomrule
  \end{tabular}
  \label{tab:op}
\end{table*}

\section{Method}
\subsection{ONNX Model Converter and Runtime}

To enhance the interoperability of pruning algorithms, we propose a universal pruning adapter named ONNXPruner, leveraging the ONNX framework. A key aspect of ONNXPruner is its ability to standardize models from diverse frameworks into the ONNX format. For instance, using PyTorch, we first load a pre-trained deep learning model and then convert it to ONNX using PyTorch's built-in conversion tool. This conversion encompasses several steps: representing the ONNX graph, replacing node operators equivalently, optimizing nodes, storing the model, checking the model structure, and verifying performance, as detailed in Table~\ref{tab:onnxcoverter}. Since the model requires fine-tuning after pruning, we avoid operator fusion and other optimizations that could alter the original structure during conversion.

After converting models from various frameworks to ONNX, the proposed ONNXPruner will automatically prune these ONNX models. For the evaluation and fine-tuning of pruned models, we use ONNX Runtime (ORT)~\cite{Onnx-runtime}, an engine that supports ONNX-based models for both inference and training through backpropagation. ORT offers efficient training and inference across different platforms and hardware, including CPUs and GPUs, by integrating with hardware-specific libraries through flexible interfaces. This capability of ORT facilitates the application and integration of diverse pruning algorithms within ONNXPruner.

\begin{figure*}[!t]
	\centering
	\begin{minipage}[b]{1\linewidth}
        \center
		\includegraphics[width=1\linewidth]{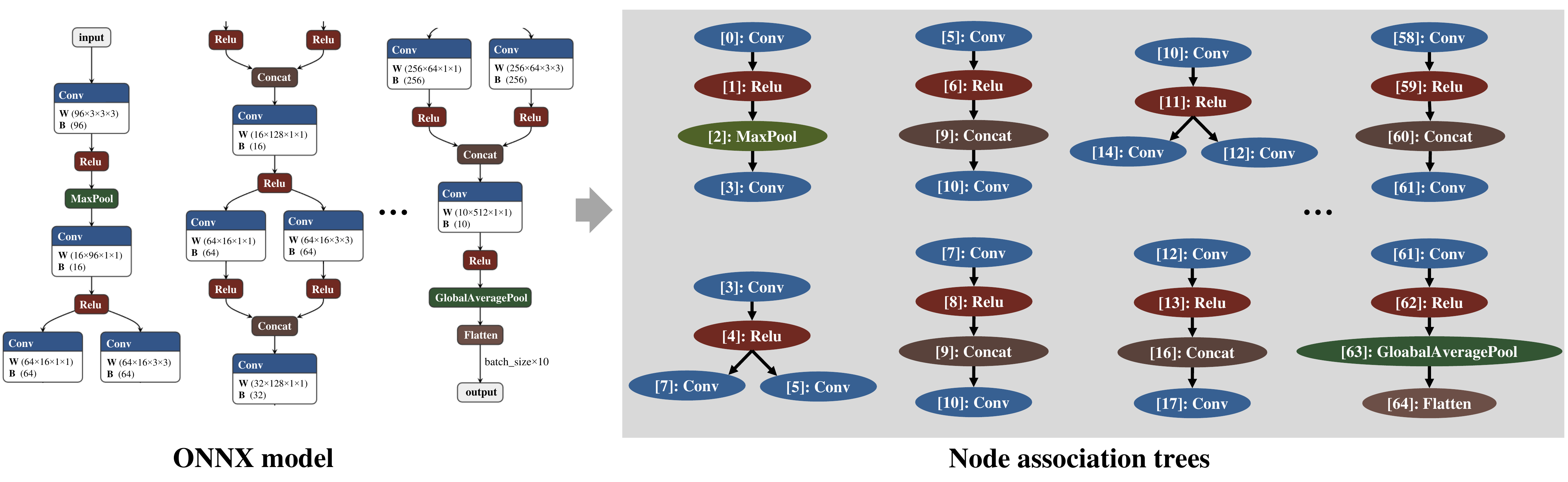}
	\end{minipage}
	\caption{An example for constructing the node association trees. We use the node graph of the ONNX model to build a node association tree for each pruned node. The attributes within these trees are assigned based on the operator type: the pruned node serves as the root, child nodes are tagged as `next' indicating further exploration, and leaf nodes are marked as `stop' indicating the end of the branch.}
	\label{fig:tree}
\end{figure*}

\subsection{Construct Node Association Tree}

In this paper, a ``pruned node" refers to a model layer requiring pruning, while ``associated nodes" are those that must be adjusted following the pruning of a pruned node, including all nodes in the sequence. To automate model pruning, identifying the associated nodes of a pruned node and understanding their relationships is crucial. Our goal is to locate all associated nodes of a pruned node and group them. To accomplish this, we initially create a node attribute library categorizing nodes into four types based on their operator roles: pruned, next-no-process, next-process, and stop-process, as detailed in Table~\ref{tab:op}. Subsequently, leveraging the node attribute library, we develop a clearer method for modeling node relationships, the \textit{node association tree}.

The construction of the node association tree  begins by navigating the ONNX node graph, starting from the pruned node as the tree's root, and identifying all nodes receiving output from this root node as their input, considered the root's children. Each child node is then evaluated based on its attribute. For next-process or next-no-process nodes, this examination continues, treating each as a new parent node. The search concludes for stop-process nodes, which are deemed leaf nodes. This approach, as summarized in Algorithm~\ref{algorithm}, facilitates the creation of node association trees for pruned nodes in any model, provided that the model's operator types are represented in the node attribute library. Figure~\ref{fig:tree} illustrates the construction of a node association tree for a pruned node in SqueezeNet~\cite{iandola2016squeezenet}.

%Finally, the root node of the node association tree is the pruned node, and all leaf nodes are stop-process nodes. 
%
%Noting that among next-process nodes include three types of nodes: self-process nodes (e.g., BN), weakly coupled nodes (e.g., Concat), and strongly coupled nodes (e.g., Sum) that need to be jointly determined with other nodes regarding the index of the pruning channel.

\begin{algorithm}[!tb]
    \caption{Construct Node Association Tree}
      \label{algorithm}
    \LinesNumbered
    \SetAlgoLined
    \KwIn{ONNX model, pruning nodels ${N[0,1...,n]}$}
    \KwOut{Node association Trees}
    \SetKwFunction{FMain}{\emph{Insertchild}}
		\SetKwProg{Fn}{Function}{}{}
		\Fn{\FMain{fathernode, childnode}}{
              fathernode $ \to $ child = childnode\\
              \Return{fathernode $ \to $ child}
		}
    \For{$i = \{0,1...,n\}$}
    {
        node = Tree(${N_i}$)\\
        \While{node not NULL}
        {
            \If{The next node of ${N_i}$ in the node graph}
            {
                \For{Next node $NN$}
                {
                    \If{$NN$}
                    {
                    node = \emph{Insertchild(node, NN)}
                    }
                    \Else
                    {
                    node = NULL
                    }
                }
            }
        }
    }
     
\end{algorithm}

\subsection{Tree-level Pruning}

The standard approach to model pruning involves pruning a pre-trained model and then fine-tuning the pruned structure to mitigate the performance loss caused by pruning. While some studies incorporate pruning at the beginning or during the training phase, the practical application of these methods is often complicated by the large dataset sizes and extended training durations. The most commonly adopted criterion for pruning in applications is the $\ell_n$-norm of the model weights~\cite{filterspruning,SFP}, a practice we adhere to develop our pruning adapter and evaluation method.

The $\ell_n$-norm technique prunes filters by comparing the $\ell_n$-norm of all channel weights in a single node, subsequently removing the corresponding input channels of the associated nodes in the following layer. However, assessing weight importance based solely on a single node's weights may not accurately reflect the true significance, especially since the parameters of the pruned node and its associated nodes are removed simultaneously. To address this issue, we put forward a \textit{tree-level pruning} method utilizing node association trees, which allows for a more accurate evaluation of which channels to prune, thus enabling the removal of less important channels more securely without adding extra components.

\begin{figure*}[!t]
	\centering
	\begin{minipage}[b]{1\linewidth}
		\includegraphics[width=1\linewidth]{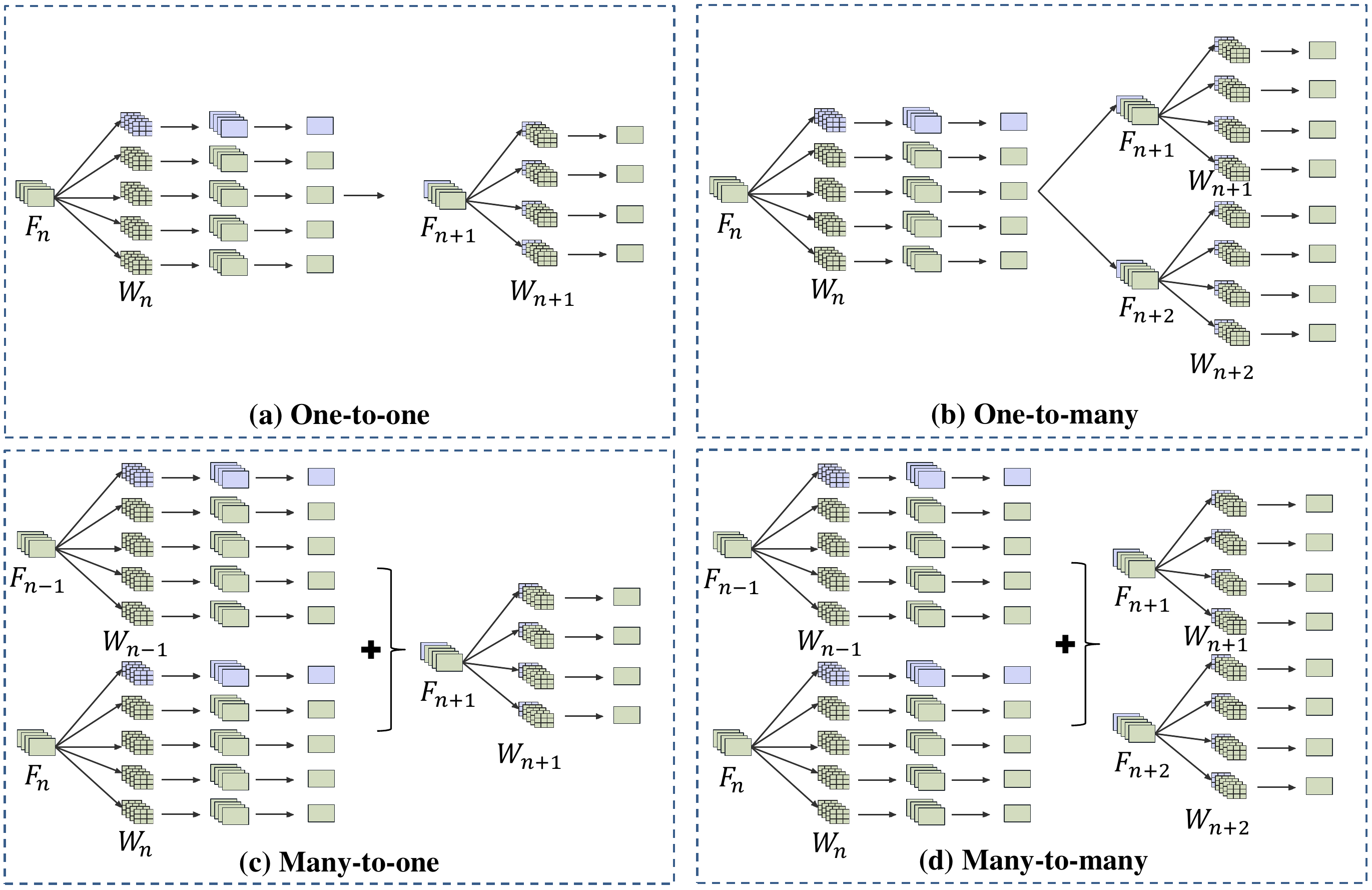}
	\end{minipage}
	\caption{We illustrate four basic configurations of pruned and associated nodes, using convolution as a representative example for simplicity. We omit associated nodes like ReLU and pooling, which do not require processing in this context. (a) One-to-one represents the standard structure in DNNs, where the output channel of one layer feeds directly into the input channel of each filter in the next layer. (b) One-to-many is prevalent in models that incorporate feature reuse, such as SqueezeNet and Inception. (c) Many-to-one typifies the conventional residual structure found in networks. (d) Many-to-many combines the characteristics of (b) and (c), representing a hybrid structure that integrates feature reuse with residual connections.}
	\label{fig:3}
\end{figure*}

We identify four types of tree-level pruning structures based on node association trees: one-to-one, one-to-many, many-to-one, and many-to-many, as shown in Figure~\ref{fig:3}.

\textbf{One-to-one connections} are fundamental in DNN models, where the output of a layer serves as the input for the next. Specifically, for input feature $F_n$, the weight ${ W_n^i }$ convolves with it, producing the $i$-th feature map for the subsequent layer. The corresponding filter in the next layer, ${ W_{n+1}^{k,i} }$,  is directly linked to ${ W_n^i }$'s output channel, as shown in Figure~\ref{fig:3}(a). 
The evaluation of a pruned node in a one-to-one structure is thus defined as:
\begin{align}%\small
{\left\| { W_n^i } \right\|_1} \times \sum_{k = 1}^j {{{\left\| { W_{n + 1}^{k,i} } \right\|}_1}},
\label{eq:1}
\end{align}
where $j$ represents the dimensional index of ${W_{n+1} }$.

\textbf{One-to-many connections} indicate a single pruned node connects to multiple associated nodes. 
For input feature $F_n$, ${ W_n^i }$ convolves with it and outputs the $ i$-th feature map affecting multiple subsequent filters, such as ${ W_{n+1}^{k,i} }$ and ${ W_{n+2}^{k,i} }$, shown in Figure~\ref{fig:3}(b). Therefore, the evaluation of the pruned node in a one-to-many structure is given by:
\begin{align}%\small
{\left\| { W_n^i } \right\|_1} \!\times\! \left( {\sum_{k = 1}^{j_1} {{{\left\| { W_{n + 1}^{k,i} } \right\|}_1}}  \!+\! \sum_{k = 1}^{j_2} {{{\left\| { W_{n + 2}^{k,i} } \right\|}_1}} } \right),
\label{eq:2}
\end{align}
where $j_1$ represents the dimensional index of ${ W_{n+1} }$ and $j_2$ represents the dimensional index of ${ W_{n+2} }$.

\textbf{Many-to-one connections}, such as in residual structures, involve combining outputs from multiple nodes into a single subsequent node. This is shown in Figure~\ref{fig:3}(c), where  ${ W_n^i }$ and ${ W_{n-1}^i }$ convolve with their respective inputs ($F_n$ and $F_{n-1}$) and their outputs are summed for the next layer's computation. The evaluation formula for a pruned node in a many-to-one structure is:
\begin{align}%\small
\left( {{{\left\| { W_{n - 1}^i } \right\|}_1} + {{\left\| { W_n^i } \right\|}_1}} \right) \times \sum_{k = 1}^j {{{\left\| { W_{n + 1}^{k,i} } \right\|}_1}},
\label{eq:3}
\end{align}
%where $j$ represents the dimensional index of ${ W_{n+1} }$.

\textbf{Many-to-many connections} combine aspects of one-to-many and many-to-one structures. The evaluation for a pruned node in such a structure is calculated as:
\begin{align}%\small
\left( {{{\left\| { W_{n \!-\! 1}^i } \right\|}_1} \!+\! {{\left\| { W_n^i } \right\|}_1}} \right) \!\times\! \left( {\sum_{k \!=\! 1}^{j_1} {{{\left\| { W_{n \!+\! 1}^{k,i} } \right\|}_1}}\!+\! \sum_{k \!=\! 1}^{j_2} {{{\left\| { W_{n \!+\! 2}^{k,i} } \right\|}_1}} } \right).
\label{eq:4}
\end{align}

In the following experimental section, we demonstrate that incorporating tree-level pruning with certain methods can yield performance on par with contemporary approaches.

% \begin{align}\small
% \begin{array}{l}
% \left( {{{\left\| { W_{n - 1}^i } \right\|}_1} + {{\left\| { W_n^i } \right\|}_1}} \right) \times \\[5mm]
% \left( {\sum_{k = 1}^{j1} {{{\left\| { W_{n + 1}^{k,i} } \right\|}_1}}+ \sum_{k = 1}^{j2} {{{\left\| { W_{n + 2}^{k,i} } \right\|}_1}} } \right).
% \end{array}
% \label{eq:4}
% \end{align}

\begin{figure}[t]
	\centering
	\begin{minipage}[b]{1\linewidth}
		\includegraphics[width=1\linewidth]{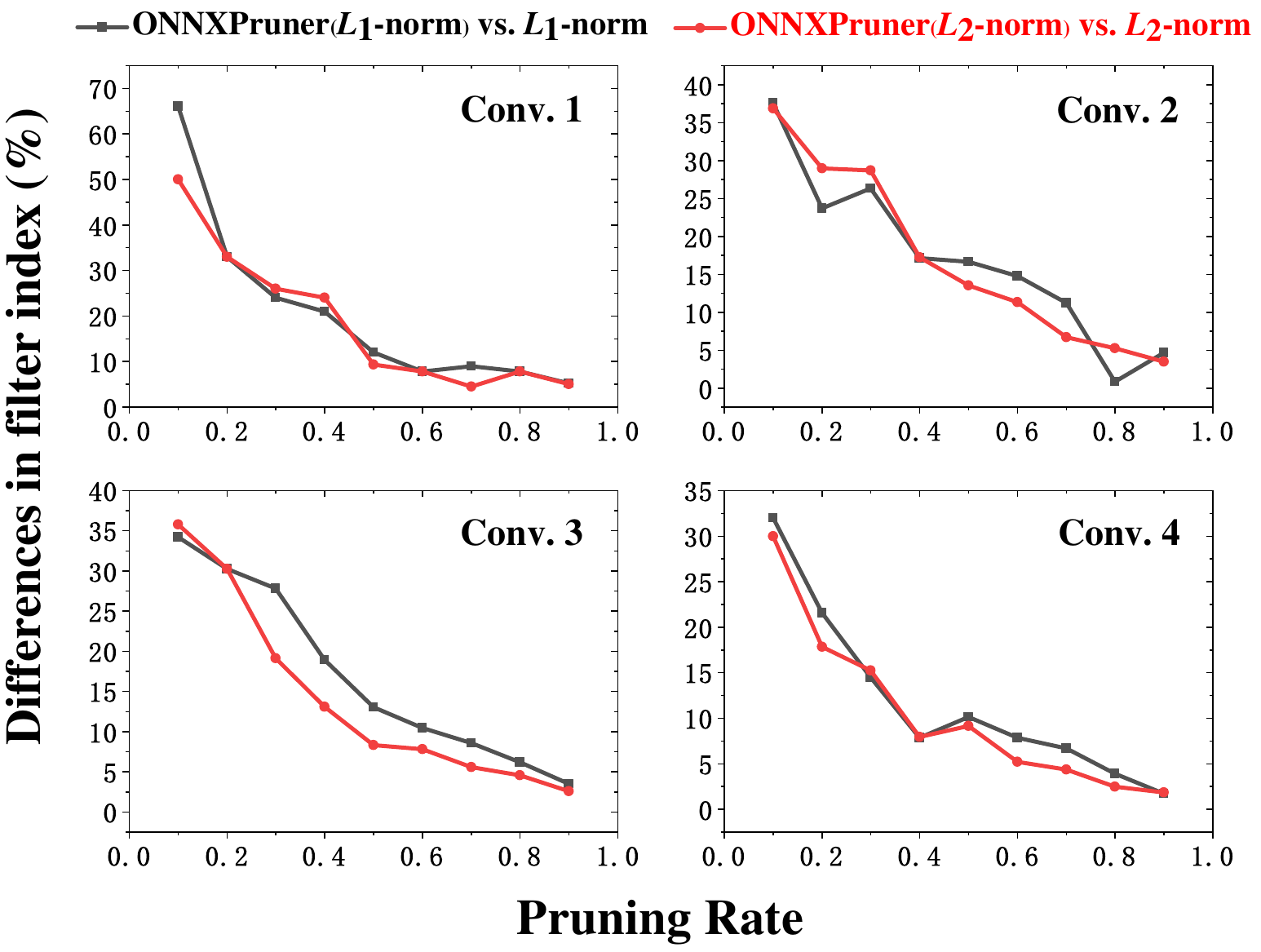}
	\end{minipage}
	\caption{Differences in filter index of ONNXPruner ($\ell_n$-norm) vs. $\ell_n$-norm.}
	\label{fig:DA}
\end{figure}

\begin{figure}[t]
	\centering
	\begin{minipage}[b]{1\linewidth}
		\includegraphics[width=1\linewidth]{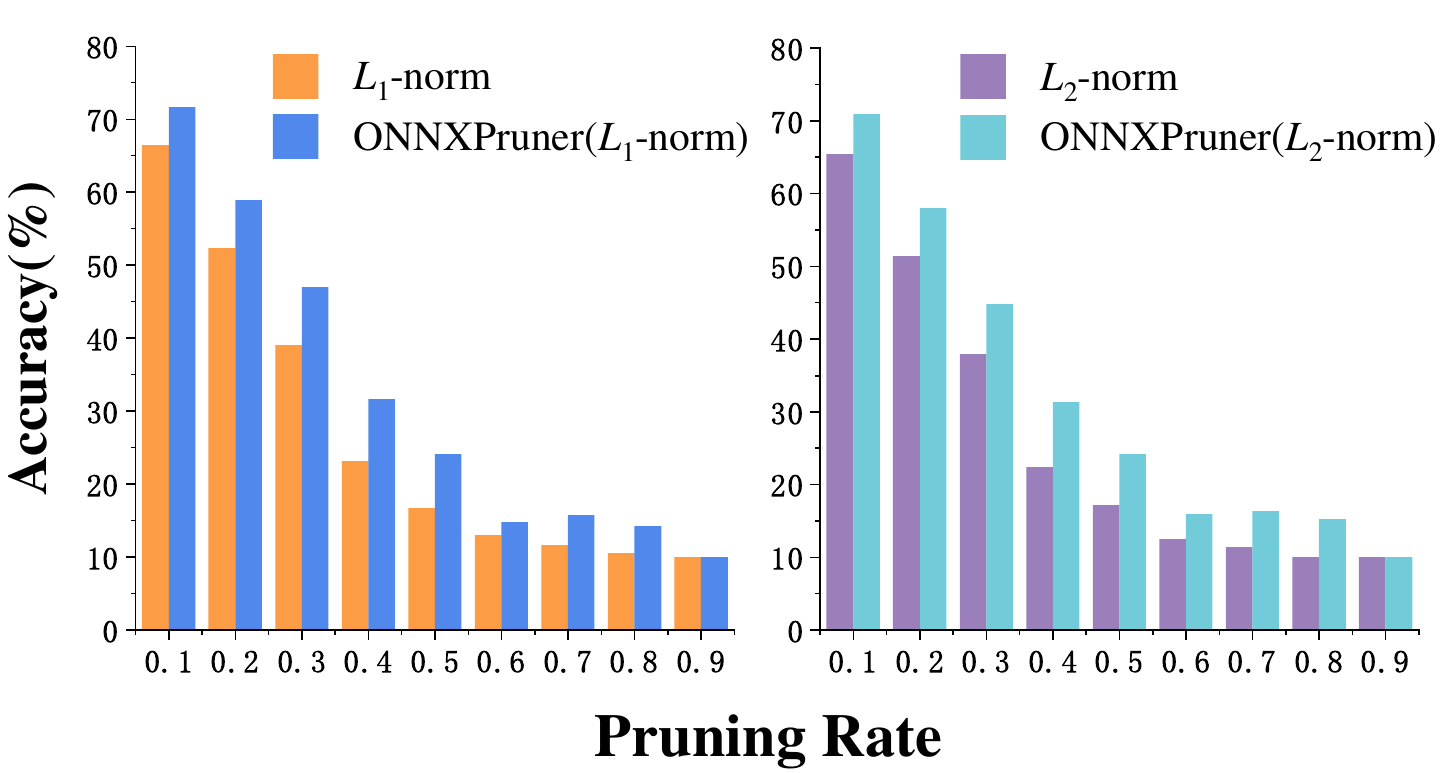}
	\end{minipage}
	\caption{Pruning of ONNXPruner ($\ell_n$-norm) and $\ell_n$-norm without fine-tuning.}
	\label{fig:DFI}
\end{figure}

\section{Experiments}

\subsection{Settings}
\textbf{Dataset.} To evaluate the performance of ONNXPruner, we utilize three well-established benchmark datasets: CIFAR-10, CIFAR-100~\cite{cifarkrizhevsky2009learning}, ImageNet~\cite{russakovsky2015imagenet} and PASCAL VOC 2012~\cite{everingham2012pascal}.  Both CIFAR-10 and CIFAR-100 comprise 60,000 images each, split into 50,000 for training and 10,000 for testing, across 10 and 100 classes, respectively. The ImageNet dataset contains 1.2 million training images and 50,000 validation images across 1,000 classes. PASCAL VOC 2012, a benchmark for semantic segmentation, includes 20 object classes plus one background class, with 1,464 training images, 1,449 validation images, and 1,456 testing images. The training set is augmented by the Semantic Boundaries Dataset~\cite{hariharan2011semantic}, totaling 10,582 training images.

\textbf{Model.} We follow the widely-used models in existing pruning studies to verify for ONNXPruner's efficacy. For CIFAR datasets, ONNXPruner is tested on various architectures, including AlexNet~\cite{krizhevsky2012imagenet}, VGG16~\cite{simonyan2014very}, VGG19~\cite{simonyan2014very}, SqueezeNet~\cite{iandola2016squeezenet}, ResNet18~\cite{he2016deep}, and ViT-B$\_$16~\cite{dosovitskiy2020image}, with each model trained independently. 
For the ImageNet dataset, we evaluate ONNXPruner on ResNet50~\cite{he2016deep}.
For the PASCAL VOC 2012 dataset, we evaluate ONNXPruner on FCN~\cite{FCNlong2015fully} and PSPNet~\cite{PSPzhao2017pyramid}.

%\textbf{Pruning and fine-tuning:} For each model, we apply pruning rate of 50\% to all convolutional and fully connected layers. 
%We use smaller learning rates and training epoch (approximately 10\% of the original training epoch) for fine-tuning.

\begin{table*}[t]%\footnotesize
    \tabcolsep=1pt
	\begin{center}
		\caption{Pruning results (classification accuracy in percentage point) for VGG16 on CIFAR-10 with different pruning rates.  Layerwise PR stands for layerwise pruning ratio. Speedup stands for the ratio of the unpruned FLOPs compared to the pruned model.}
		\begin{spacing}{1.3}
			\begin{tabular}{p{5cm}<{}p{2.5cm}<{}p{2.5cm}<{}p{2.5cm}<{}p{2.5cm}<{}}
				\Xhline{1.5px}
                \multicolumn{5}{c}{Unpruned acc. 92.64, Params: 14.16M, FLOPs: 0.58G}\\
                \hline
				Layerwise PR & 0.3 & 0.5 & 0.7 & 0.9\\
                Sparsity & 49.89\% & 74.02\% & 90.19\% & 98.62\%\\
                Speedup & 1.89$\times$ & 3.37$\times$ & 7.57$\times$ & 30.11$\times$\\
				\hline%\hline
				$\ell_1$-norm~\cite{filterspruning} & 91.34 & 90.11 & 88.77 & 84.58\\
                ONNXPruner ($\ell_1$) & 92.29 {\tiny \color{blue}{($\uparrow$0.95)}} & 91.07  {\tiny \color{blue}{($\uparrow$0.96)}} & 90.16  {\tiny \color{blue}{($\uparrow$1.39)}} & 85.26  {\tiny \color{blue}{($\uparrow$0.68)}}\\
                \hdashline
                $\ell_2$-norm~\cite{SFP} & 91.34 & 90.16 & 88.75  & 84.58\\
                ONNXPruner ($\ell_2$) & 92.26 {\tiny \color{blue}{($\uparrow$0.92)}} & 91.00 {\tiny \color{blue}{($\uparrow$0.84)}} & 90.15 {\tiny \color{blue}{($\uparrow$1.40)}}  & 85.17 {\tiny \color{blue}{($\uparrow$0.59)}}\\
                
                \hdashline
                Hrank$^*$~\cite{Hranklin2020hrank} & 91.52 & 90.47 & 88.91  & 84.51\\
                ONNXPruner (Hrank$^*$) & 92.33 {\tiny \color{blue}{($\uparrow$0.81)}} & 91.50 {\tiny \color{blue}{($\uparrow$1.03)}} & 90.19 {\tiny \color{blue}{($\uparrow$1.28)}}  & 85.20 {\tiny \color{blue}{($\uparrow$0.69)}}\\
				\Xhline{1.5px}
			\end{tabular}
			\label{tab:dpr}
		\end{spacing}
	\end{center}
\end{table*}

\subsection{Effectiveness of Tree-level Pruning}

The proposed tree-level pruning method evaluates the importance of filters by considering both pruned and associated nodes, which distinguishes it from evaluations that focus on single nodes~\cite{filterspruning,SFP,Hranklin2020hrank}. This approach is illustrated in Figure~\ref{fig:DA}, showing the difference in filter selection between ONNXPruner (using $\ell_1$-norm and $\ell_2$-norm) and the traditional $\ell_1$-norm or $\ell_2$-norm methods, with AlexNet~\cite{krizhevsky2012imagenet} as the example. The difference in filter indices between the methods is calculated as follows: 
\begin{equation}
    \frac{{\text{index}(\text{ONNXPruner} ({\ell_n}\text{-norm})) \cap \text{index}({\ell_n}\text{-norm})}}{{\text{Count}(\text{index}({\ell_n}\text{-norm}))}}.
\end{equation}

Results in Figure~\ref{fig:DA} indicate a notable discrepancy in filter indices between ONNXPruner ($\ell_n$-norm) and traditional methods across various layers, especially at lower pruning rates (below 0.3), where differences exceed 20\%. This gap diminishes as the pruning rate increases, mainly because of the broader inclusion of pruned channels.

To assess the efficacy of tree-level pruning without the influence of finetuning, we test the accuracy of models pruned with ONNXPruner, as shown in Figure~\ref{fig:DFI}. Using CIFAR-10 as the dataset and pruning rates ranging from 0.1 to 0.9, we observe that ONNXPruner generally enhances accuracy across various pruning levels. Notably, at a pruning rate of 0.9, accuracy approaches that of random guessing, reflecting the extensive loss of model capacity due to excessive pruning.

\subsection{Results on CIFAR}

First, we validate the performance of ONNXPruner across various pruning rates on the VGG16 model using the CIFAR-10 dataset, as shown in Table~\ref{tab:dpr}. We set specific layerwise pruning rates for all convolutional and fully connected layers, where a rate of 0.3 implies removing 30\% of the parameters in these layers. The pruned models are then fine-tuned for 10 epochs using ONNXRuntime with an initial learning rate of $10^{-3}$. We incorporate baseline algorithms such as $\ell_1$-norm~\cite{filterspruning}, $\ell_2$-norm~\cite{SFP}, and Hrank~\cite{Hranklin2020hrank} into ONNXPruner for tree-level filter importance evaluation, with Hrank$^*$ indicating a combined approach of using feature map rank for convolutional layers and $\ell_1$-norm for fully connected layers. The results demonstrate that the proposed ONNXPruner, without any additional constraints, can integrate various weight evaluation algorithms and outperform the baselines at various pruning rates.

To further evaluate the performance of ONNXPruner on different models, we conduct additional tests with a fixed pruning rate of 0.5, detailed in Table~\ref{tab:cifar}. Comparing ONNXPruner against other methods like Taylor-FO~\cite{r1_molchanov2019importance}, GReg-2~\cite{r2_wang2020neural}, reimpl.~($\ell_1$-norm)~\cite{r3_wang2023state}, TPP~\cite{r4_wang2022trainability} and DepGraph~($\ell_1$-norm)~\cite{fang2023depgraph}, which focus on optimizing filter importance evaluation but require extra components or constraints, ONNXPruner shows superior or comparable performance. The ``-" indicates models incompatible with certain algorithms, thus not yielding results.

\begin{table*}[t]%\footnotesize
    \tabcolsep=1pt
	\begin{center}
		\caption{Pruning results (classification accuracy in percentage point) for various models on CIFAR-10 and CIFAR-100. We pruning 50\% of the parameters in all convolutional layers and fully connected layers on models AlexNet~\cite{krizhevsky2012imagenet}, SqueezeNet~\cite{iandola2016squeezenet}, VGG16~\cite{simonyan2014very}, VGG19~\cite{simonyan2014very}, ResNet18~\cite{he2016deep}, and ViT-B$\_$16~\cite{dosovitskiy2020image}. Taylor-FO~\cite{r1_molchanov2019importance}, GReg-2~\cite{r2_wang2020neural}, reimpl. ($\ell_1$-norm)~\cite{r3_wang2023state}, TPP~\cite{r4_wang2022trainability}, DepGraph~\cite{fang2023depgraph} are existing methods that perform well in optimizing the evaluation strategies. Since these methods require additional components or operations, we implemented these on Pytorch and adapted them to different model structures, including residual connections. We implemented the $\ell_1$-norm~\cite{filterspruning}, $\ell_2$-norm~\cite{SFP}, and Hrank~\cite{Hranklin2020hrank} methods on the ONNX format and embedded them into ONNXPruner. The blue color indicates the performance change after ONNXPruner embeds and optimizes these methods.}
		\begin{spacing}{1.3}
			\begin{tabular}{p{1.5cm}<{\centering}p{4cm}<{}p{2cm}<{}p{2cm}<{}p{2cm}<{}p{2cm}<{}p{2cm}<{}p{2cm}<{}}
				\Xhline{1.5px}
				Data & Method & AlexNet  & SqueezeNet & VGG16  & VGG19  & ResNet18 & ViT-B$\_$16\\
                 \hline 
                \multirow{12}{*}{\rotatebox[origin=c]{90}{CIFAR-10}}&Unpruned& 79.27  & 83.15 & 92.64  & 92.58  & 93.02 & 98.67\\
                \cline{2-8}
                &Taylor-FO~\cite{r1_molchanov2019importance} & 77.65 & 80.01 & 90.70  & 90.63 &  90.91 & 96.02\\
                &GReg-2~\cite{r2_wang2020neural} & 77.98 & 80.69 & 91.01  & 91.03 &  91.13 & --\\
                &reimpl. ($\ell_1$-norm)~\cite{r3_wang2023state} & 78.09 & 80.71 & 91.08  & 91.24 &  91.35 & 96.17\\
                &TPP~\cite{r4_wang2022trainability} & 78.19 & 81.93 & 91.47  & \textbf{91.89} &  \textbf{91.72} & --\\
                &DepGraph ($\ell_1$-norm)~\cite{fang2023depgraph} & 78.01 & 81.86 & 91.29  & 91.64 &  91.38 & 96.62\\
    
				\cline{2-8} 
				&$\ell_1$-norm~\cite{filterspruning} & 77.08 & 79.32 & 90.11  & 90.21 &  90.38 & 95.94\\
                &ONNXPruner ($\ell_1$) & 78.13 {\tiny \color{blue}{($\uparrow$1.05)}} & 81.76 {\tiny \color{blue}{($\uparrow$2.44)}} & 91.07 {\tiny \color{blue}{($\uparrow$0.96)}}  & 91.77 {\tiny \color{blue}{($\uparrow$1.56)}} &  91.68 {\tiny \color{blue}{($\uparrow$1.30)}} & \textbf{96.82} {\tiny \color{blue}{($\uparrow$0.88)}} \\
				\cdashline{2-8}
                &$\ell_2$-norm~\cite{SFP} & 77.01 & 79.29 & 90.16  & 90.51 &  90.44 & 95.31 \\
                &ONNXPruner ($\ell_2$) & 78.06 {\tiny \color{blue}{($\uparrow$1.05)}} & 81.78 {\tiny \color{blue}{($\uparrow$2.49)}}& 91.00 {\tiny \color{blue}{($\uparrow$0.84)}} & 91.91 {\tiny \color{blue}{($\uparrow$1.40)}} &  91.73 {\tiny \color{blue}{($\uparrow$1.29)}} & 96.14 {\tiny \color{blue}{($\uparrow$0.83)}} \\
				\cdashline{2-8}
                &Hrank$^*$~\cite{Hranklin2020hrank} & 77.33 & 79.57 & 90.47  & 90.59 &  90.40 & --\\
                &ONNXPruner (Hrank$^*$) & \textbf{78.20} {\tiny \color{blue}{($\uparrow$0.87)}} & \textbf{82.00} {\tiny \color{blue}{($\uparrow$2.43)}} & \textbf{91.50} {\tiny \color{blue}{($\uparrow$1.03)}} & 91.87 {\tiny \color{blue}{($\uparrow$1.28)}} &  91.61 {\tiny \color{blue}{($\uparrow$1.21)}} & --\\
                
                \cline{1-8}
                \multirow{12}{*}{\rotatebox[origin=c]{90}{CIFAR-100}}&Unpruned& 55.41 & 69.41 & 72.93 & 72.94 & 75.21  & 90.97\\
                \cline{2-8}
                &Taylor-FO~\cite{r1_molchanov2019importance} & 52.34 & 66.32 & 69.52  & 70.24 &  72.04 & 85.80\\
                &GReg-2~\cite{r2_wang2020neural} & 52.47 & 66.51 & 69.70  & 70.49 &  72.17 & --\\
                &reimpl. ($\ell_1$-norm)~\cite{r3_wang2023state} & 52.89 & 66.79 & 70.29  & 70.88 &  72.69 & 86.02\\
                &TPP~\cite{r4_wang2022trainability} & 53.20 & \textbf{67.00} & 70.44  & 71.33 &  \textbf{73.78} & --\\
                &DepGraph ($\ell_1$-norm)~\cite{fang2023depgraph} & 53.14 & 66.83 & 70.37 & 71.22 &  73.16 & 86.47\\
                \cline{2-8}
                &$\ell_1$-norm~\cite{filterspruning} & 52.03 & 66.01 & 69.35  & 70.01 &  71.80 & 85.41\\
                &ONNXPruner ($\ell_1$) & 53.17 {\tiny \color{blue}{($\uparrow$1.14)}} & 66.70 {\tiny \color{blue}{($\uparrow$0.69)}} & 70.43 {\tiny \color{blue}{($\uparrow$1.08)}} & 71.23 {\tiny \color{blue}{($\uparrow$1.22)}} &  73.01 {\tiny \color{blue}{($\uparrow$1.21)}} & \textbf{86.90} {\tiny \color{blue}{($\uparrow$1.49)}}\\
				\cdashline{2-8}
                &$\ell_2$-norm~\cite{SFP} & 52.09 & 66.08 & 69.31  & 68.98 &  71.91 & 85.02\\
                &ONNXPruner ($\ell_2$) & 53.16 {\tiny \color{blue}{($\uparrow$1.07)}} & 66.74 {\tiny \color{blue}{($\uparrow$0.66)}} & 70.35 {\tiny \color{blue}{($\uparrow$1.04)}} & 71.07 {\tiny \color{blue}{($\uparrow$1.09)}} &  73.15 {\tiny \color{blue}{($\uparrow$1.24)}}& 86.77 {\tiny \color{blue}{($\uparrow$1.75)}}\\
				\cdashline{2-8}
                &Hrank$^*$~\cite{Hranklin2020hrank} & 52.19 & 66.35 & 69.78  & 70.26 &  72.37& --\\
                &ONNXPruner (Hrank$^*$) & \textbf{53.29} {\tiny \color{blue}{($\uparrow$1.10)}} & 66.91 {\tiny \color{blue}{($\uparrow$0.56)}} & \textbf{70.50} {\tiny \color{blue}{($\uparrow$0.72)}} & \textbf{71.66} {\tiny \color{blue}{($\uparrow$1.40)}} &  73.70 {\tiny \color{blue}{($\uparrow$1.33)}} &--\\
				\Xhline{1.5px}
			\end{tabular}
			\label{tab:cifar}
		 \end{spacing}
	\end{center}
 \vspace{-0.8cm}
\end{table*}

We also explore the adaptability of ONNXPruner by integrating it with during-training pruning methods like FPGM~\cite{he2019filter} and SFP~\cite{SFP}. As shown in Table~\ref{tab:cifar_pruning_druning_training}, ONNXPruner not only accommodates these methods but enhances their optimization performance. ONNXPruner's design facilitates easy integration with existing and new pruning methods by simply updating the criteria for importance evaluation. Additionally, the overhead for constructing the node association tree and performing tree-level model evaluation on a model like ResNet is approximately 1.93 seconds on an RTX 3090 GPU, a minimal delay compared to the overall training duration of about 2.2 hours.

\subsection{Results on ImageNet}

To evaluate ONNXPruner on a larger dataset, we test its performance and efficiency on ImageNet for ResNet50. 
We show several pruning methods, including $\ell_1$-norm~\cite{filterspruning}, Hrank~\cite{Hranklin2020hrank}, FPGM~\cite{he2019filter}, and SFP~\cite{SFP}, and their performance when integrated into ONNXPruner.
The results, as presented in Table~\ref{tab:imagenet}, show that ONNXPruner outperforms each pruning method by achieving a higher Pruned Top-1 accuracy with a smaller decrease in Top-1 accuracy, all while maintaining consistent speed improvements.

\subsection{Results on PASCAL VOC 2012}

Beyond image classification, we extend the application of ONNXPruner to the task of image segmentation, testing it on two widely-used models: FCN~\cite{FCNlong2015fully} and PSPNet~\cite{PSPzhao2017pyramid}. The evaluation, using mean Intersection over Union (mean IoU) as the metric and presented in Table~\ref{tab:VOC}, involves baseline methods such as $\ell_1$-norm~\cite{filterspruning}, $\ell_2$-norm~\cite{SFP}, and Hrank~\cite{Hranklin2020hrank} integrated within ONNXPruner. Adapting these methods to FCN and PSPNet manually is a time-consuming process, but the proposed ONNXPruner simplifies this, allowing for quick adaptation to model structures for pruning. The use of a tree-level pruning strategy enables more accurate filter importance assessment. The mean IoU results in Table~\ref{tab:VOC} indicate that ONNXPruner delivers superior performance compared to baseline methods.

\begin{table*}[t]%\footnotesize
    \tabcolsep=1pt
	\begin{center}
		\caption{Pruning-during-training results (classification accuracy in percentage point) for various models on CIFAR-10. }
		\begin{spacing}{1.3}
			\begin{tabular}{p{4cm}<{}p{2.5cm}<{\centering}p{2.5cm}<{\centering}p{2.5cm}<{\centering}p{2.5cm}<{\centering}p{2.5cm}<{\centering}}
				\Xhline{1.5px}
				Method & AlexNet& SqueezeNet & VGG16  & ResNet18\\
                 \hline
				FPGM~\cite{he2019filter} & 77.08 & 79.32 & 90.11  &  90.38\\
                ONNXPruner (FPGM) & \textbf{78.13} & \textbf{81.76}  & \textbf{91.07}  &  \textbf{91.68} \\
                \cdashline{1-6}
                SFP~\cite{SFP} & 77.01 & 79.29 & 90.16  &  90.44 \\
                ONNXPruner (SFP) & \textbf{78.06} & \textbf{81.78} & \textbf{91.00} &  \textbf{91.73} \\
				\Xhline{1.5px}
			\end{tabular}
			\label{tab:cifar_pruning_druning_training}
		 \end{spacing}
	\end{center}
 %\vspace{-0.8cm}
\end{table*}

\vspace{3cm}

\begin{table}[h]%\footnotesize
    \tabcolsep=1pt
	\begin{center}
		\caption{Pruning results (classification accuracy in percentage point) for ResNet50 on ImageNet.}
		\begin{spacing}{1.3}
			\begin{tabular}{p{3.3cm}<{}p{1.9cm}<{\centering}p{1.8cm}<{\centering}p{1.5cm}<{\centering}}
				\Xhline{1.5px}
				Method & Pruned top-1& Top-1 drop & Speedup\\
                 \hline
                $\ell_1$-norm~\cite{filterspruning} &74.23 & 1.90 & 1.78$\times$\\
                ONNXPruner ($\ell_1$) & \textbf{74.89} & \textbf{1.24}  & \textbf{1.78$\times$}\\
                \cdashline{1-4}
                Hrank~\cite{Hranklin2020hrank} & 74.98 & 1.15 & 1.78$\times$\\
                ONNXPruner (Hrank) & \textbf{75.44} & \textbf{0.69} & \textbf{1.78$\times$}\\
                \cdashline{1-4}
				FPGM~\cite{he2019filter} & 75.01 & 1.12 & 1.59$\times$\\
                ONNXPruner (FPGM) & \textbf{75.43} & \textbf{0.70}  & \textbf{1.59$\times$}\\
                \cdashline{1-4}
                SFP~\cite{SFP} & 74.61 & 1.52 & 1.76$\times$\\
                ONNXPruner (SFP) &\textbf{75.22} & \textbf{0.91} & \textbf{1.76$\times$}\\
				\Xhline{1.5px}
			\end{tabular}
			\label{tab:imagenet}
		 \end{spacing}
	\end{center}
 %\vspace{-0.9cm}
\end{table}

\begin{table}[h]%\footnotesize
    \tabcolsep=1pt
	\begin{center}
		\caption{Pruning results (mean IoU) for various models on PASCAL VOC 2012.} %We pruning 50\% of the parameters in all convolutional layers and fully connected layers on  FCN~\cite{FCNlong2015fully} and PSPNet~\cite{PSPzhao2017pyramid}.}
		\begin{spacing}{1.3}
			\begin{tabular}{p{3.5cm}<{}p{2.25cm}<{}p{2.25cm}<{}}
				\Xhline{1.5px}
				Method & FCN~\cite{FCNlong2015fully} & PSPNet~\cite{PSPzhao2017pyramid} \\ %first cell is occupied by the multirow
                \hline
                Unpruned & 62.2 & 82.6\\
                \hline				
				$\ell_1$-norm~\cite{filterspruning} & 60.3 & 78.7\\
                ONNXPruner ($\ell_1$) & 61.0 {\tiny \color{blue}{($\uparrow$0.7)}} & 80.1 {\tiny \color{blue}{($\uparrow$1.4)}}\\
				\hdashline
                $\ell_2$-norm~\cite{SFP} & 60.3 & 78.6\\
                ONNXPruner ($\ell_2$) & 61.1 {\tiny \color{blue}{($\uparrow$0.8)}} & 79.8 {\tiny \color{blue}{($\uparrow$1.2)}}\\
				\hdashline
                Hrank$^*$~\cite{Hranklin2020hrank} & 60.7 & 78.7\\
                ONNXPruner (Hrank$^*$) & 61.6 {\tiny \color{blue}{($\uparrow$0.9)}} & 80.1 {\tiny \color{blue}{($\uparrow$1.4)}}\\
				\Xhline{1.5px}
			\end{tabular}
			\label{tab:VOC}
		\end{spacing}
	\end{center}
 %\vspace{-0.8cm}
\end{table}

\section{Conclusion}\vspace{-0.1cm}
This paper proposes ONNXPruner, a general model pruning adapter for ONNX models. It automates the application of pruning algorithms across different model structures, aiding developers in enhancing practical applications. Future developments for ONNXPruner will focus on expanding support for more operator types, accommodating more model architectures like Bidirectional LSTM/GRU, and further enhancing the efficiency of tree-level pruning.

\bibliographystyle{IEEEtran}
\bibliography{cite}

\begin{thebibliography}{10}
\providecommand{\url}[1]{#1}
\csname url@samestyle\endcsname
\providecommand{\newblock}{\relax}
\providecommand{\bibinfo}[2]{#2}
\providecommand{\BIBentrySTDinterwordspacing}{\spaceskip=0pt\relax}
\providecommand{\BIBentryALTinterwordstretchfactor}{4}
\providecommand{\BIBentryALTinterwordspacing}{\spaceskip=\fontdimen2\font plus
\BIBentryALTinterwordstretchfactor\fontdimen3\font minus \fontdimen4\font\relax}
\providecommand{\BIBforeignlanguage}[2]{{%
\expandafter\ifx\csname l@#1\endcsname\relax
\typeout{** WARNING: IEEEtran.bst: No hyphenation pattern has been}%
\typeout{** loaded for the language `#1'. Using the pattern for}%
\typeout{** the default language instead.}%
\else
\language=\csname l@#1\endcsname
\fi
#2}}
\providecommand{\BIBdecl}{\relax}
\BIBdecl

\bibitem{fang2023depgraph}
G.~Fang, X.~Ma, M.~Song, M.~B. Mi, and X.~Wang, ``Depgraph: Towards any structural pruning,'' in \emph{IEEE Conf. Comput. Vis. Pattern Recog.}, 2023, pp. 16\,091--16\,101.

\bibitem{han2015learning}
S.~Han, J.~Pool, J.~Tran, and W.~Dally, ``Learning both weights and connections for efficient neural network,'' \emph{Adv. Neural Inform. Process. Syst.}, vol.~28, 2015.

\bibitem{liu2021group}
L.~Liu, S.~Zhang, Z.~Kuang, A.~Zhou, J.-H. Xue, X.~Wang, Y.~Chen, W.~Yang, Q.~Liao, and W.~Zhang, ``Group fisher pruning for practical network compression,'' in \emph{International Conference on Machine Learning}.\hskip 1em plus 0.5em minus 0.4em\relax PMLR, 2021, pp. 7021--7032.

\bibitem{jing2021meta}
Y.~Jing, Y.~Yang, X.~Wang, M.~Song, and D.~Tao, ``Meta-aggregator: Learning to aggregate for 1-bit graph neural networks,'' in \emph{IEEE Conf. Comput. Vis. Pattern Recog.}, 2021, pp. 5301--5310.

\bibitem{kuchaiev2019nemo}
O.~Kuchaiev, J.~Li, H.~Nguyen, O.~Hrinchuk, R.~Leary, B.~Ginsburg, S.~Kriman, S.~Beliaev, V.~Lavrukhin, J.~Cook \emph{et~al.}, ``Nemo: a toolkit for building ai applications using neural modules,'' \emph{arXiv preprint arXiv:1909.09577}, 2019.

\bibitem{yao2021wenet}
Z.~Yao, D.~Wu, X.~Wang, B.~Zhang, F.~Yu, C.~Yang, Z.~Peng, X.~Chen, L.~Xie, and X.~Lei, ``Wenet: Production oriented streaming and non-streaming end-to-end speech recognition toolkit,'' \emph{arXiv preprint arXiv:2102.01547}, 2021.

\bibitem{8416559}
J.-H. Luo, H.~Zhang, H.-Y. Zhou, C.-W. Xie, J.~Wu, and W.~Lin, ``Thinet: Pruning cnn filters for a thinner net,'' \emph{IEEE Transactions on Pattern Analysis and Machine Intelligence}, vol.~41, no.~10, pp. 2525--2538, 2019.

\bibitem{9457173}
W.~Niu, Z.~Li, X.~Ma, P.~Dong, G.~Zhou, X.~Qian, X.~Lin, Y.~Wang, and B.~Ren, ``Grim: A general, real-time deep learning inference framework for mobile devices based on fine-grained structured weight sparsity,'' \emph{IEEE Transactions on Pattern Analysis and Machine Intelligence}, vol.~44, no.~10, pp. 6224--6239, 2022.

\bibitem{wang2021convolutional}
Z.~Wang, C.~Li, and X.~Wang, ``Convolutional neural network pruning with structural redundancy reduction,'' in \emph{IEEE Conf. Comput. Vis. Pattern Recog.}, 2021, pp. 14\,913--14\,922.

\bibitem{chao2020directional}
S.-K. Chao, Z.~Wang, Y.~Xing, and G.~Cheng, ``Directional pruning of deep neural networks,'' \emph{Adv. Neural Inform. Process. Syst.}, vol.~33, pp. 13\,986--13\,998, 2020.

\bibitem{10330640}
Y.~He and L.~Xiao, ``Structured pruning for deep convolutional neural networks: A survey,'' \emph{IEEE Transactions on Pattern Analysis and Machine Intelligence}, vol.~46, no.~5, pp. 2900--2919, 2024.

\bibitem{kwon2020structured}
S.~J. Kwon, D.~Lee, B.~Kim, P.~Kapoor, B.~Park, and G.-Y. Wei, ``Structured compression by weight encryption for unstructured pruning and quantization,'' in \emph{IEEE Conf. Comput. Vis. Pattern Recog.}, 2020, pp. 1909--1918.

\bibitem{chen2021orthant}
T.~Chen, T.~Ding, B.~Ji, G.~Wang, Y.~Shi, J.~Tian, S.~Yi, X.~Tu, and Z.~Zhu, ``Orthant based proximal stochastic gradient method for l1-regularized optimization,'' in \emph{Machine Learning and Knowledge Discovery in Databases}, 2021, pp. 57--73.

\bibitem{chen2020neural}
T.~Chen, B.~Ji, Y.~Shi, T.~Ding, B.~Fang, S.~Yi, and X.~Tu, ``Neural network compression via sparse optimization,'' \emph{arXiv preprint arXiv:2011.04868}, 2020.

\bibitem{liu2017learning}
Z.~Liu, J.~Li, Z.~Shen, G.~Huang, S.~Yan, and C.~Zhang, ``Learning efficient convolutional networks through network slimming,'' in \emph{Int. Conf. Comput. Vis.}, 2017, pp. 2736--2744.

\bibitem{li2016pruning}
H.~Li, A.~Kadav, I.~Durdanovic, H.~Samet, and H.~P. Graf, ``Pruning filters for efficient convnets,'' \emph{arXiv preprint arXiv:1608.08710}, 2016.

\bibitem{he2019filter}
Y.~He, P.~Liu, Z.~Wang, Z.~Hu, and Y.~Yang, ``Filter pruning via geometric median for deep convolutional neural networks acceleration,'' in \emph{IEEE Conf. Comput. Vis. Pattern Recog.}, 2019, pp. 4340--4349.

\bibitem{malach2020proving}
E.~Malach, G.~Yehudai, S.~Shalev-Schwartz, and O.~Shamir, ``Proving the lottery ticket hypothesis: Pruning is all you need,'' in \emph{International Conference on Machine Learning}.\hskip 1em plus 0.5em minus 0.4em\relax PMLR, 2020, pp. 6682--6691.

\bibitem{chen2021only}
T.~Chen, B.~Ji, T.~Ding, B.~Fang, G.~Wang, Z.~Zhu, L.~Liang, Y.~Shi, S.~Yi, and X.~Tu, ``Only train once: A one-shot neural network training and pruning framework,'' \emph{Adv. Neural Inform. Process. Syst.}, pp. 19\,637--19\,651, 2021.

\bibitem{chen2023otov}
T.~Chen, L.~Liang, T.~Ding, Z.~Zhu, and I.~Zharkov, ``{OTO}v2: Automatic, generic, user-friendly,'' in \emph{Int. Conf. Learn. Represent.}, 2023.

\bibitem{liu2021content}
Y.~Liu, Z.~Shu, Y.~Li, Z.~Lin, F.~Perazzi, and S.-Y. Kung, ``Content-aware gan compression,'' in \emph{IEEE Conf. Comput. Vis. Pattern Recog.}, 2021, pp. 12\,156--12\,166.

\bibitem{ONNX}
``Onnx,'' in \emph{https://onnx.ai/}, 2019.

\bibitem{you2019gate}
Z.~You, K.~Yan, J.~Ye, M.~Ma, and P.~Wang, ``Gate decorator: Global filter pruning method for accelerating deep convolutional neural networks,'' \emph{Adv. Neural Inform. Process. Syst.}, vol.~32, 2019.

\bibitem{filterspruning}
H.~Li, A.~Kadav, I.~Durdanovic, H.~Samet, and H.~P. Graf, ``Pruning filters for efficient convnets,'' in \emph{Int. Conf. Learn. Represent.}, 2017, pp. 24--26.

\bibitem{SFP}
Y.~He, G.~Kang, X.~Dong, Y.~Fu, and Y.~Yang, ``Soft filter pruning for accelerating deep convolutional neural networks,'' in \emph{IJCAI}, 2018, pp. 2234--2240.

\bibitem{Hranklin2020hrank}
M.~Lin, R.~Ji, Y.~Wang, Y.~Zhang, B.~Zhang, Y.~Tian, and L.~Shao, ``Hrank: Filter pruning using high-rank feature map,'' in \emph{IEEE Conf. Comput. Vis. Pattern Recog.}, 2020, pp. 1529--1538.

\bibitem{krizhevsky2012imagenet}
A.~Krizhevsky, I.~Sutskever, and G.~E. Hinton, ``Imagenet classification with deep convolutional neural networks,'' \emph{Adv. Neural Inform. Process. Syst.}, vol.~25, 2012.

\bibitem{iandola2016squeezenet}
F.~N. Iandola, S.~Han, M.~W. Moskewicz, K.~Ashraf, W.~J. Dally, and K.~Keutzer, ``Squeezenet: Alexnet-level accuracy with 50x fewer parameters and< 0.5 mb model size,'' \emph{arXiv preprint arXiv:1602.07360}, 2016.

\bibitem{simonyan2014very}
K.~Simonyan and A.~Zisserman, ``Very deep convolutional networks for large-scale image recognition,'' \emph{arXiv preprint arXiv:1409.1556}, 2014.

\bibitem{he2016deep}
K.~He, X.~Zhang, S.~Ren, and J.~Sun, ``Deep residual learning for image recognition,'' in \emph{IEEE Conf. Comput. Vis. Pattern Recog.}, 2016, pp. 770--778.

\bibitem{FCNlong2015fully}
J.~Long, E.~Shelhamer, and T.~Darrell, ``Fully convolutional networks for semantic segmentation,'' in \emph{IEEE Conf. Comput. Vis. Pattern Recog.}, 2015, pp. 3431--3440.

\bibitem{PSPzhao2017pyramid}
H.~Zhao, J.~Shi, X.~Qi, X.~Wang, and J.~Jia, ``Pyramid scene parsing network,'' in \emph{IEEE Conf. Comput. Vis. Pattern Recog.}, 2017, pp. 2881--2890.

\bibitem{dosovitskiy2020image}
A.~Dosovitskiy, L.~Beyer, A.~Kolesnikov, D.~Weissenborn, X.~Zhai, T.~Unterthiner, M.~Dehghani, M.~Minderer, G.~Heigold, S.~Gelly \emph{et~al.}, ``An image is worth 16x16 words: Transformers for image recognition at scale,'' \emph{arXiv preprint arXiv:2010.11929}, 2020.

\bibitem{jajal2023analysis}
P.~Jajal, W.~Jiang, A.~Tewari, J.~Woo, Y.-H. Lu, G.~K. Thiruvathukal, and J.~C. Davis, ``Analysis of failures and risks in deep learning model converters: A case study in the onnx ecosystem,'' \emph{arXiv preprint arXiv:2303.17708}, 2023.

\bibitem{paszke2019pytorch}
A.~Paszke, S.~Gross, F.~Massa, A.~Lerer, J.~Bradbury, G.~Chanan, T.~Killeen, Z.~Lin, N.~Gimelshein, L.~Antiga \emph{et~al.}, ``Pytorch: An imperative style, high-performance deep learning library,'' \emph{Adv. Neural Inform. Process. Syst.}, vol.~32, 2019.

\bibitem{abadi2016tensorflow}
M.~Abadi, P.~Barham, J.~Chen, Z.~Chen, A.~Davis, J.~Dean, M.~Devin, S.~Ghemawat, G.~Irving, M.~Isard \emph{et~al.}, ``$\{$TensorFlow$\}$: a system for $\{$Large-Scale$\}$ machine learning,'' in \emph{operating systems design and implementation}, 2016, pp. 265--283.

\bibitem{chen2015mxnet}
T.~Chen, M.~Li, Y.~Li, M.~Lin, N.~Wang, M.~Wang, T.~Xiao, B.~Xu, C.~Zhang, and Z.~Zhang, ``Mxnet: A flexible and efficient machine learning library for heterogeneous distributed systems,'' \emph{arXiv preprint arXiv:1512.01274}, 2015.

\bibitem{chen2018tvm}
T.~Chen, T.~Moreau, Z.~Jiang, L.~Zheng, E.~Yan, H.~Shen, M.~Cowan, L.~Wang, Y.~Hu, L.~Ceze \emph{et~al.}, ``$\{$TVM$\}$: An automated $\{$End-to-End$\}$ optimizing compiler for deep learning,'' in \emph{13th USENIX Symposium on Operating Systems Design and Implementation (OSDI 18)}, 2018, pp. 578--594.

\bibitem{TensorRT}
``Tensorrt developer documentation,'' in \emph{https://docs.nvidia. com/deeplearning/tensorrt/developer-guide/index.html}.

\bibitem{OpenVINO}
``Openvino documentation,'' in \emph{https://docs.openvino.ai/ latest/home.html}, 2021.

\bibitem{8485719}
S.~Chen and Q.~Zhao, ``Shallowing deep networks: Layer-wise pruning based on feature representations,'' \emph{IEEE Transactions on Pattern Analysis and Machine Intelligence}, vol.~41, no.~12, pp. 3048--3056, 2019.

\bibitem{dong2017learning}
X.~Dong, S.~Chen, and S.~Pan, ``Learning to prune deep neural networks via layer-wise optimal brain surgeon,'' \emph{Adv. Neural Inform. Process. Syst.}, vol.~30, 2017.

\bibitem{lee2019signal}
N.~Lee, T.~Ajanthan, S.~Gould, and P.~H. Torr, ``A signal propagation perspective for pruning neural networks at initialization,'' \emph{arXiv preprint arXiv:1906.06307}, 2019.

\bibitem{park2020lookahead}
S.~Park, J.~Lee, S.~Mo, and J.~Shin, ``Lookahead: A far-sighted alternative of magnitude-based pruning,'' \emph{arXiv preprint arXiv:2002.04809}, 2020.

\bibitem{Namhoon:ICLR2019}
N.~Lee, T.~Ajanthan, and P.~H.~S. Torr, ``Snip: single-shot network pruning based on connection sensitivity,'' in \emph{Int. Conf. Learn. Represent.}, 2019.

\bibitem{Chaoqi:ICLR2020}
C.~Wang, G.~Zhang, and R.~B. Grosse, ``Picking winning tickets before training by preserving gradient flow,'' in \emph{Int. Conf. Learn. Represent.}, 2020.

\bibitem{Pau:ICLR2021}
P.~de~Jorge, A.~Sanyal, H.~S. Behl, P.~H.~S. Torr, G.~Rogez, and P.~K. Dokania, ``Progressive skeletonization: Trimming more fat from a network at initialization,'' in \emph{Int. Conf. Learn. Represent.}, 2021.

\bibitem{Jonathan:ICML2020}
J.~Frankle, G.~K. Dziugaite, D.~M. Roy, and M.~Carbin, ``Linear mode connectivity and the lottery ticket hypothesis,'' in \emph{International Conference on Machine Learning}, 2020, pp. 3259--3269.

\bibitem{Yang:IJCAI2018}
Y.~He, G.~Kang, X.~Dong, Y.~Fu, and Y.~Yang, ``Soft filter pruning for accelerating deep convolutional neural networks,'' in \emph{IJCAI}, 2018, pp. 2234--2240.

\bibitem{Tao:ICLR2020}
T.~Lin, S.~U. Stich, L.~Barba, D.~Dmitriev, and M.~Jaggi, ``Dynamic model pruning with feedback,'' in \emph{Int. Conf. Learn. Represent.}, 2020.

\bibitem{10265172}
J.~Mu, M.~Wang, F.~Zhu, J.~Yang, W.~Lin, and W.~Zhang, ``Boosting the convergence of reinforcement learning-based auto-pruning using historical data,'' \emph{IEEE Transactions on Computer-Aided Design of Integrated Circuits and Systems}, vol.~43, no.~2, pp. 548--561, 2024.

\bibitem{Yuchen:CVPR2021}
Y.~Liu, Z.~Shu, Y.~Li, Z.~Lin, F.~Perazzi, and S.~Kung, ``Content-aware {GAN} compression,'' in \emph{IEEE Conf. Comput. Vis. Pattern Recog.}, 2021, pp. 12\,156--12\,166.

\bibitem{Lewei:CVPR2021}
L.~Yao, R.~Pi, H.~Xu, W.~Zhang, Z.~Li, and T.~Zhang, ``Joint-detnas: Upgrade your detector with nas, pruning and dynamic distillation,'' in \emph{IEEE Conf. Comput. Vis. Pattern Recog.}, 2021, pp. 10\,175--10\,184.

\bibitem{huang2017densely}
G.~Huang, Z.~Liu, L.~Van Der~Maaten, and K.~Q. Weinberger, ``Densely connected convolutional networks,'' in \emph{Int. Conf. Comput. Vis.}, 2017, pp. 4700--4708.

\bibitem{he2017channel}
Y.~He, X.~Zhang, and J.~Sun, ``Channel pruning for accelerating very deep neural networks,'' in \emph{Int. Conf. Comput. Vis.}, 2017, pp. 1389--1397.

\bibitem{wang2024structurally}
X.~Wang, J.~Rachwan, S.~G{\"u}nnemann, and B.~Charpentier, ``Structurally prune anything: Any architecture, any framework, any time,'' \emph{arXiv preprint arXiv:2403.18955}, 2024.

\bibitem{Onnx-runtime}
``Onnx-runtime environment,'' in \emph{https://onnxruntime.ai/}, 2021.

\bibitem{cifarkrizhevsky2009learning}
A.~Krizhevsky, G.~Hinton \emph{et~al.}, ``Learning multiple layers of features from tiny images,'' 2009.

\bibitem{russakovsky2015imagenet}
O.~Russakovsky, J.~Deng, H.~Su, J.~Krause, S.~Satheesh, S.~Ma, Z.~Huang, A.~Karpathy, A.~Khosla, M.~Bernstein \emph{et~al.}, ``Imagenet large scale visual recognition challenge,'' \emph{International journal of computer vision}, pp. 211--252, 2015.

\bibitem{everingham2012pascal}
M.~Everingham and J.~Winn, ``The pascal visual object classes challenge 2012 (voc2012) development kit,'' \emph{Pattern Anal. Stat. Model. Comput. Learn., Tech. Rep}, no. 1-45, 2012.

\bibitem{hariharan2011semantic}
B.~Hariharan, P.~Arbel{\'a}ez, L.~Bourdev, S.~Maji, and J.~Malik, ``Semantic contours from inverse detectors,'' in \emph{Int. Conf. Comput. Vis.}, 2011, pp. 991--998.

\bibitem{r1_molchanov2019importance}
P.~Molchanov, A.~Mallya, S.~Tyree, I.~Frosio, and J.~Kautz, ``Importance estimation for neural network pruning,'' in \emph{IEEE Conf. Comput. Vis. Pattern Recog.}, 2019, pp. 11\,264--11\,272.

\bibitem{r2_wang2020neural}
H.~Wang, C.~Qin, Y.~Zhang, and Y.~Fu, ``Neural pruning via growing regularization,'' \emph{arXiv preprint arXiv:2012.09243}, 2020.

\bibitem{r3_wang2023state}
H.~Wang, C.~Qin, Y.~Bai, and Y.~Fu, ``Why is the state of neural network pruning so confusing? on the fairness, comparison setup, and trainability in network pruning,'' \emph{arXiv preprint arXiv:2301.05219}, 2023.

\bibitem{r4_wang2022trainability}
H.~Wang and Y.~Fu, ``Trainability preserving neural structured pruning,'' \emph{arXiv preprint arXiv:2207.12534}, 2022.

\end{thebibliography}

\end{document}